\pdfoutput=1

\documentclass[11pt]{article}

\usepackage{EMNLP2022}

\usepackage{times}
\usepackage{latexsym}

\usepackage[T1]{fontenc}

\usepackage[utf8]{inputenc}

\usepackage{graphicx}
\graphicspath{ {./images/} }

\usepackage{microtype}

\usepackage{inconsolata}
\usepackage{amsmath}

\title{PromptShots at the FinNLP-2022 ERAI Tasks:\\ Pairwise Comparison and Unsupervised Ranking}

\author{Peratham Wiriyathammabhum \\
  \texttt{peratham.bkk@gmail.com} \\
  \vspace{-3\baselineskip}
  }

\begin{document}
\maketitle
\begin{abstract}
This report describes our PromptShots submissions to a shared task on Evaluating the Rationales of Amateur Investors (ERAI). We participated in both pairwise comparison and unsupervised ranking tasks. For pairwise comparison, we employed instruction-based models based on T5-small and OpenAI InstructGPT language models. Surprisingly, we observed OpenAI InstructGPT language model few-shot trained on Chinese data works best in our submissions, ranking $3^{rd}$ on the maximal loss (ML) pairwise accuracy. This model works better than training on the Google translated English data by a large margin, where the English few-shot trained InstructGPT model even performs worse than an instruction-based T5-small model finetuned on the English data. However, all instruction-based submissions do not perform well on the maximal potential profit (MPP) pairwise accuracy where there are more data and learning signals. The Chinese few-shot trained InstructGPT model still performs best in our setting. For unsupervised ranking, we utilized many language models, including many financial-specific ones, and Bayesian lexicons unsupervised-learned on both Chinese and English words using a method-of-moments estimator. All our submissions rank best in the MPP ranking, from $1^{st}$ to $3^{rd}$. However, they all do not perform well for ML scoring. Therefore, both MPP and ML scores need different treatments since we treated MPP and ML using the same formula. Our only difference is the treatment of market sentiment lexicons.
\end{abstract}

\section{Introduction}
Evaluating the rationals of amateur investors (ERAI) \cite{chen2021evaluating, chen2021opinion} is a shared task on evaluating social media opinions on the topic of investments and whether they are going to be useful or not. Mining high-quality opinions by inspecting their supporting rationales might utilize the wisdom of the crowd on social media. Previous work \cite{chen2021evaluating} proposes stylistic and semantic features to filter out noisy crowd opinions which may not be high-quality and profitable. There are two settings in this ERAI shared task, pairwise comparison and unsupervised ranking. These settings sort out the opinions based on two metrics, higher maximal potential profit (MPP) and lower maximal loss (ML). In pairwise comparison, two posts are given with a binary label whether the MPP and ML of the first post are more or less than the second post. In unsupervised ranking, the goal is to filter and keep the top 10\% posts based on MPP and ML given a set of unranked posts.

For pairwise comparison, our best submission ranks $3^{rd}$ on the maximal loss (ML) pairwise accuracy on the leaderboard\footnote{\url{https://sites.google.com/nlg.csie.ntu.edu.tw/finnlp-2022-emnlp/erai-shared-task}}. For unsupervised ranking, our best submission ranks $1^{st}$ on the maximal potential profit (MPP) ranking. The codes for our systems are open-sourced and available at our GitHub repository\footnote{\url{https://github.com/perathambkk/finnlp_erai_shared_task_emnlp2022}}.
\section{Models}
\subsection{Pairwise Comparison}
For pairwise comparison, we utilized instruction-based models based on T5-small \cite{raffel2020exploring} and OpenAI InstructGPT language models \cite{ouyang2022training} in a few-shot prompt-based setting \cite{brown2020language}.

\subsubsection{T5}
T5 is an encoder-decoder language model which was trained by treating every text processing problem as a ``text-to-text" problem to unify NLP tasks using only a single model, loss function, hyperparameter set, etc. The input texts will be encoded and the T5 decoder will decode them. Specifically, T5 was unsupervised-pretrained by denoising masked inputs on the ``Colossal Clean Crawled Corpus" (C4) dataset, Common Crawled from web scraping. Then, T5 can be further supervised finetuned using the ``text-to-text'' format and T5's decoder will be in the teacher forcing mode where the decoder will be trained using input and a right-shifted target sequence. T5 architecture is pretty much the same as the vanilla transformer \cite{vaswani2017attention} except removing the LayerNorm bias term, placing the LayerNorm outside the residual path, and using a relative position embedding \cite{shaw-etal-2018-self}. 

In the T5 paper, the authors state that T5 can be specified which task it should perform by adding a task-specific textual prefix to the original input sequence before feeding into the model. Therefore, we take T5-small as an instruction-based model using the input prompt template, `post1 : \%s post2 : \%s  </s>', where the \%s contains texts from the corresponding post, and the output prompt, `maximal potential profit (MPP) : \%s maximal loss (ML) : \%s </s>', where the \%s contains the MPP and the ML corresponding labels accordingly. This is similar to the baseline system in the FLUTE figurative language understanding dataset paper \cite{chakrabarty2022flute}, however, in our case, the T5-small is expected to jointly predict both MPP and ML in one forward pass of preparing the probability tensor. We use top-p sampling for text generation \cite{holtzman2019curious}.

\subsubsection{OpenAI's InstructGPT}
The OpenAI API has many variants of InstructGPT language models based on the GPT-3 autoregressive language model to conveniently perform various NLP tasks with the prompt library. The InstructGPT was trained with a human-in-the-loop style and is claimed to be better at following instructions, more truthful, and less toxic than the GPT-3. In this shared task, we engineered the prompts for InstructGPT using a few-shot learning setting, as in the GPT-3 paper \cite{NEURIPS2020_1457c0d6}, where few data instances were given from the target task/domain. Each data instance will become a prompt as `post1 : d[`post1'] post2: d[`post2'] > maximal potential profit (MPP)| \%s\# maximal loss (ML)| \%s.', where d[`post1'] and d[`post2'] are texts from the corresponding post. Then, we append the query we want to predict MPP and ML as just a truncated template, `post1 : d[`post1'] post2: d[`post2'] >', and let the language model generate the rest. 

We use the `text-davinci-002' model and randomly construct those few-shot prompts where each prompt will be a length of around $4,000$ because of the API token length limit. We use the same setting and the model pipeline for both of our submissions $2$ and $3$ where we use the Chinese posts as d[`post1'] and d[`post2'] for our submission $2$ and the Google-translated English posts for our submission $3$. By this we mean, for example, we use the same tokenizer for Chinese and English. Therefore, these systems are very simple and to-go prompt-based systems. We had done very minimal parameter tuning to the model, only prompt engineering. For a survey in prompt-based systems, please consider \cite{liu2021pre}. 

\subsection{Unsupervised Ranking}
For unsupervised ranking, we utilized many financial and general language models and Bayesian lexicons in both Chinese and English.

\subsubsection{Base Model}
Our first submission, our base model, consists of a stylistic length feature \cite{zong-etal-2020-measuring} derived from the opinion (sub)word lengths segmented using the `hfl/chinese-bert-wwm-ext' tokenizer \cite{cui2021pre}, prediction scores from FinBERT-FLS \cite{huang2020finbert}, a professional lexicon count from FinProLex \cite{chen2021evaluating}, and a market sentiment lexicon count from NTUSD-Fin \cite{chen2018ntusd}.

In the measuring forecasting skill from text paper \cite{zong-etal-2020-measuring}, the authors observe various linguistics phenomena indicating that skilled forecasters tend to write significantly longer justifications because of more rationale. For example, skilled forecasters also provide less readability, because of the usage of more complex languages, and less emotion, because of the usage of less emotional languages as neutral sentiments. Moreover, skilled forecasters tend to use more cardinal numbers, prepositions, and nouns. They tend to use fewer verbs and pronouns. Therefore, in this base model, we just stick with the lengths of justifications as our simplest skill indicator. 

FinBERT \cite{huang2020finbert} is essentially BERT \cite{devlin-etal-2019-bert} customized for financial texts, pretrained on corporate filings, analyst reports, and earnings conference call transcripts, which differ from normal texts in both vocabulary and writing style. In the FinBERT paper, FinBERT outperforms all other methods, Loughran McDonald lexicon, and machine learning algorithms, especially in negative financial sentiment prediction, when finetuned for financial sentiment analysis (FinBERT-tone). FinBERT was finetuned in two additional tasks, labeling environment, social, and governance (ESG) discussions and labeling forward-looking statements (FLS), from firms’ corporate social responsibility (CSR) reports and management discussion and analyses (MD\&As) textual sentences. For this base model, we sum and normalize the prediction logit outputs from FinBERT-FLS classes, $\{FLS, NON\_FLS, NOT\_FLS\}$ on each sentence of the textual inputs as our scores. 

FinProLex \cite{chen2021evaluating} is a Chinese financial lexicon derived from Bloomberg Terminal and PTT Stock Taiwanese social media platform, containing $5,162$ tokens from professional analysts' reports and social media posts paired with expertise scores. FinProLex uses Point-wise Mutual Information (PMI), as in \cite{turney-2002-thumbs, li-shah-2017-learning}, to measure the association strengths between a word and either the positive or negative lexicon. The formula of the expert-like score (ELScore) of a given word $w$ is as follows:
\begin{multline}
    ELScore_w = PMI(w, analyst) \\
    - PMI(w, amateur),
\end{multline}
\begin{multline}
   ELScore_w =\log_{2}\frac{p(w, analyst)}{p(w)p(analyst)} \\- \log_{2}\frac{p(w, amateur)}{p(w)p(amateur)},
\end{multline}
where $analyst$ and $amateur$ are labels of whether a given word is from an analyst report or an amateur post. This is the difference value between the PMI scores measuring how much a term is associated with either analyst or amateur documents. This is similar to the term's sentiment score ($S_{PMI}$) \cite{li-shah-2017-learning} which is
\begin{equation}\label{eq:spmi}
S_{PMI} = PMI(w, bullish) - PMI(w, bearish).
\end{equation}

FinProLex tends to include hard words, complex semantics, noun phrase modifiers, content words, transition words, personal pronouns, and negative words as experts tend to use most of them, except personal pronouns and negative words which are used more by amateurs, based on the paper findings that can be summarized as experts tend to evaluate pricing and valuations while amateurs tend to predict the stock movements. 

NTUSD-Fin is an English lexicon for market sentiment analysis from StockTwits, containing $8,331$ words, $112$ hashtags, and $115$ emojis. We used their market sentiment scores which are also computed essentially from equation (\ref{eq:spmi}).

We aggregated the scores to predict MPP and ML using these heuristic functions ($\operatorname{base-1}$),
\begin{multline}    
MPP = len + FinProLex + |FinWord>0| \\+ (FLS + 0.5  \operatorname{NON\_FLS} - \operatorname{NOT\_FLS}),
\end{multline}
\begin{multline}
ML = len + FinProLex + |FinWord<0| \\+ (FLS + 0.5 \operatorname{NON\_FLS} - \operatorname{NOT\_FLS}).
\end{multline}

We simply used a weighted sum as our heuristic function. We grouped similar scores together. $len + FinProLex$ are stylistic features where we put an equal weight of $1$ for each of them. We used $FinWord$ as a switch feature for either MPP or ML that would behave differently because of the market sentiment based on our belief. MPP posts should be from a bullish market while ML posts should instead be from a bearish market. $FLS$ has $3$ different class scores so we weighted $1$ for a positive class, $0.5$ for a less positive one, and $-1$ for a negative class. The weights are just our rule-of-thumb (make-up numbers that we felt they made sense solely from our intuitions). 

It is like trying to intuitively come up with a good feature weighting number for a Maximum Entropy (MaxEnt) model. From our heuristic functions, we just down-weighted some scores and specify some negative interactions. We did not normalize the weighting into probabilities but the ranking should be the same anyway. Most weightings are uniformly the same number.
\begin{itemize}
\item If a score should positively correlate with the target, we should give a high weight. 
\item If a score should weakly correlate with the target, we should give a low weight. 
\item If a score should negatively correlate with the target, we should give a high negative weight.
\item For the rest that we are not certain of, they should retain a maximum entropy (uniformity).
\item These heuristics can be estimated with intuitions and give an intuitive unsupervised aggregated scoring function. 
\end{itemize}
 However, we submitted the same function for MPP and ML to get a sense of using the same strategy for both bullish and bearish markets. 

\subsubsection{Bayesian Lexicons}
Next, we added Bayesian lexicons \cite{eisenstein2017unsupervised} (by fitting FinProLex and NTUSD-Fin), FinBERT-tone \cite{huang2020finbert}, (fitted) Loughran-McDonald financial sentiment lexicon (LM) \cite{loughran2011liability} and Part-of-Speech (POS) features \cite{zong-etal-2020-measuring} into the score aggregators. We would like to note that these lexicons are not multi-word (only unigrams) so they are not expected to be able to handle negations except the creators of those lexicons had made them handle some kind of negations, like in the LM lexicon Fin-Pos list. The authors use bigram to quadgrams counts when that bigram to quadgram follows some negation patterns. Our second and third submissions differ in the normalization of scores and Bayesian lexicon variants.

Loughran-McDonald financial sentiment lexicon (LM) \cite{loughran2011liability} was created because the Harvard Psychosociological Dictionary, specifically, the Harvard-IV-4 TagNeg (H4N) file, does not perform well in financial and accounting domains. Lots of Harvard dictionary negative words are not negative in finance.

Bayesian lexicon learns predictive weights for each word in a lexicon using a method-of-moments estimator from co-occurrence statistics without any labels as a special case of multinomial Na\"ive Bayes. For the second submission, we use the Dirichlet Compound Multinomial likelihood to reduce effective counts for repetitive words. For the third submission, we use the multinomial likelihood model. For example, when we fitted the LM lexicon using the pairwise comparison data, we gave $0.02626$ to `good', $0.00501$ to `optimistic', and $0.00278$ to `highest'. For LM negative words, we gave $0.00243$ to `decline', $0.00234$ to sharply, and $0.00186$ to `difficult'.

Our POS features are motivated by the measuring forecasting skill from text paper. We simply counted cardinal numbers, nouns, and verbs from Chinese jieba segmented texts. Then, these counts were normalized into the range of $[0, 1]$. 

For these submissions, we sum and normalize the prediction logit outputs from FinBERT-tone classes, $\{pos\_tone, neg\_tone\}$ on each sentence of the textual inputs as our scores. 

For the second submission, we aggregated the scores to predict MPP and ML using these heuristic functions ($\operatorname{bayesdcm-2}$),
\begin{multline}    
MPP = len + FinProLex + |FinWord>0| \\ + (FLS + 0.5 \operatorname{NON\_FLS} - \operatorname{NOT\_FLS}) \\+ (pos\_tone - neg\_tone + LM) \\+ (nouns + cards - verbs),
\end{multline}
\begin{multline}
ML = len + FinProLex + |FinWord<0| \\+ (FLS + 0.5 \operatorname{NON\_FLS} - \operatorname{NOT\_FLS}) \\+ (pos\_tone - neg\_tone + LM) \\+ (nouns + cards - verbs).
\end{multline}

For the third submission, we aggregated the scores to predict MPP and ML using these heuristic functions ($\operatorname{multinomial-3}$),
\begin{multline}    
MPP = 0.5(len + FinProLex) \\+ 0.33(FLS + 0.5 \operatorname{NON\_FLS} - \operatorname{NOT\_FLS})\\+ 0.33(pos\_tone - neg\_tone + LM) \\+ 0.33(nouns + cards - verbs)\\+ |FinWord>0|,
\end{multline}
\begin{multline}
ML = 0.5(len + FinProLex) \\+ 0.33(FLS + 0.5 \operatorname{NON\_FLS} - \operatorname{NOT\_FLS})\\+ 0.33(pos\_tone - neg\_tone + LM) \\+ 0.33(nouns + cards - verbs)\\+ |FinWord<0|.
\end{multline}
In these functions, we tried to group and reweigh the scores as normalization. If two or more scores mean the same thing, we might double count. 
\section{Experimental Results}
In our experiments, most of our submissions (except T5-small) are intuition-based heuristics, and we did not even measure neither their training nor validation performance at all during the competition. We did not use any data augmentation techniques. 
\begin{table}[t] 
\centering
\caption{MPP and ML accuracies of our models in pairwise comparison test data. (The numbers in subscript are submission rankings on the leaderboard. The symbol $\dagger$ denotes a top-3 performance.)}
\vspace{-0.5\baselineskip}
\label{st1table}
\begin{tabular}{|l|c|c|}
\hline
 Model    &  MPP acc. & ML acc.\\
\hline
T5-small  & $47.13_{16}$ & $45.98_{10}$ \\
InstructGPT-zh  & $\mathbf{48.28_{14}}$ & $\mathbf{54.02_3}\dagger$ \\
InstructGPT-en  & $47.13_{16}$ & $41.38_{13}$\\
\hline
FinNLP-22 best & $62.07$& $59.77$ \\
\hline
\end{tabular}
\end{table}

\begin{table}[t] 
\centering
\caption{Average MPP and ML from top $10\%$ posts of our models in unsupervised ranking test data. (The numbers in subscript are submission rankings on the leaderboard. The symbol $\dagger$ denotes a top-3 performance and the symbol $\ddagger$ denotes the score beats the baseline.)}
\vspace{-0.5\baselineskip}
\label{st2table}
\begin{tabular}{|l|c|c|}
\hline
 Model    &  avg. MPP & avg. ML \\
 \hline
Stylistic baseline & $17.61\%$& $-2.46\%$ \\
\hline
base-1  & $22.53\%_{3}\dagger\ddagger$ & $\mathbf{-7.80\%_{11}}$ \\
bayesdcm-2  & $\mathbf{24.39\%_{1}\dagger\ddagger}$ & $-13.04\%_{16}$ \\
multinomial-3  & $23.76\%_{2}\dagger\ddagger$ & $-12.33\%_{15}$\\
\hline
FinNLP-22 best & $24.39\%$ (ours)& $-2.46\%$ \\
\hline
\end{tabular}
\vspace{-1\baselineskip}
\end{table}

\subsection{Pairwise Comparison}
The experimental results in Table.\ref{st1table} show that the OpenAI InstructGPT language model few-shot trained on Chinese data works best in our submissions, ranking 3rd on the maximal loss (ML) pairwise accuracy, even better than instead training on the Google translated English data by a large margin, where the English few-shot trained InstructGPT model even performs worse than an instruction-based T5-small model finetuned on the English data. However, all instruction-based submissions do not perform well on the maximal potential profit (MPP) pairwise accuracy where there are more data and learning signals, nonetheless, the Chinese few-shot trained InstructGPT model still performs best in our setting.

We additionally split the training data into a held-out train/val split and evaluated our methods on the val split in Table \ref{st3table}. The results are a bit different since the English version of the InstructGPT works better. However, we did not hope for an accurate cross-validation estimation given a small amount of data. Using leave-one-out validation (LOOCV) or $k$-fold cross-validation with a high value of $k$ can produce a better estimation but they are costly. We might be able to generate more data pairs, but we decided to keep the same setting. 

\begin{table}[t] 
\centering
\caption{Additional experiments on using our pairwise comparison methods on a held-out train/val split (ratio=$0.3$). The evaluation metric is accuracy.}
\vspace{-0.5\baselineskip}
\label{st3table}
\begin{tabular}{|l|c|c|}
\hline
 Model    &   MPP acc. &  ML acc.\\
\hline
T5-small  & 0.4833 & \textbf{0.6000}\\
InstructGPT-zh  & 0.4667 & 0.4167\\
InstructGPT-en  & \textbf{0.6167} & 0.4667 \\
\hline
\end{tabular}
\vspace{-1\baselineskip}
\end{table}

\subsection{Unsupervised Ranking}
For the unsupervised ranking task, we utilized many language models, including many financial-specific ones, and Bayesian lexicons unsupervisely learned on both Chinese and English words. All of our submissions rank best in the MPP ranking, from 1st to 3rd in this task. However, they all do not perform well for the ML scoring. Therefore, both MPP and ML scores need different treatments substantially since we treated MPP and ML using the same formula. Our only difference is the treatment of market sentiment lexicons. We feel that the sentiment features, or mostly semantic features, might be negatively correlated, weakly correlated, or even uncorrelated with ML because the stylistic baseline performs best, and our base submission performs better than our Bayesian lexicon submissions. 

We conducted additional experiments on unsupervised ranking by using the whole training set of the pairwise comparison data. We compared all posts using our scoring functions in Table \Ref{st4table}. The results show not much difference among our methods. When we tried to evaluate using the pairwise comparison accuracy, the results show no difference ($0.545$ MPP comparison acc. and $0.525$ ML comparison acc.) as our methods were not designed for that.

\begin{table}[t] 
\centering
\caption{Additional experiments on using our unsupervised ranking methods to rank all posts of the pairwise data. The evaluation metrics are average MPP and average ML of the top $10\%$ posts.}
\vspace{-0.5\baselineskip}
\label{st4table}
\begin{tabular}{|l|c|c|}
\hline
 Model    &  avg. MPP & avg. ML\\
\hline
base-1  & 0.2083 & -0.2108\\
bayesdcm-2  & \textbf{0.2085} & \textbf{-0.2104} \\
multinomial-3  & \textbf{0.2085} & \textbf{-0.2104} \\
\hline
\end{tabular}
\vspace{-1\baselineskip}
\end{table}
\section{Conclusion}
This report describes our systems for a shared task of evaluating the rationales of amateur investors at FinNLP-2022. From the experimental results in pairwise comparison, we conclude that few-shot prompted instruction-based language models can work reasonably well in low resource settings with minimal training efforts but might need quite accurate data from sources since using translated data seems not to perform well. From the experimental results in unsupervised ranking, financial language models perform well and Bayesian-fitting the lexicons helps improve the performance. Also, the heuristic function design needs to differ between MPP and ML. 

\section*{Limitations}
We only sampled a relatively small portion of models and draw conclusions. We also conducted experiments only on one dataset for evaluating the rationales of amateur investors. Besides, the dataset is in Chinese with English translation using Google Translate. Lots of our methods rely on the translated data. 

Because we are limited to only three submissions, we don't know how each feature set contributes to the score. There were no ablations. However, the shared task organizers released the test data with ground truths in private. 

The authors are self-affiliated and do not represent any entities. The authors also participated in the shared task under many severe unattended local personal criminal events in their home countries. There might be some unintentional errors and physical limitations based on these unlawful interruptions. Even at the time of drafting this report, the authors suffer from unknown toxin flumes spraying into their places. We want to participate in the shared task because it is fun and educational. We apologize for any errors in this report. We tried our best.

\section*{Ethics Statement}
Scientific work published at EMNLP 2022 must comply with the \href{https://www.aclweb.org/portal/content/acl-code-ethics}{ACL Ethics Policy}. We, the authors, hope the intended uses of our systems are for peace, well-being, and social good only. No harm. 

\section*{Acknowledgments}
We would like to thank anonymous reviewers for their constructive feedback and additional experiment suggestions. 

\bibliography{anthology,custom}
\bibliographystyle{acl_natbib}

\end{document}